# AN APPROACH FOR ROBOTS TO DEAL WITH OBJECTS

Sidiq S. Hidayat[1,3], Bong Keung Kim[2], Kohtaro Ohba[2]

[1]Dep. of Intelligent Interaction Technology, University of Tsukuba, Tsukuba, JAPAN
[2]Intelligent Systems Institute, The National Institute of Advanced Industrial Science & Technology (AIST) Tsukuba, Tsukuba, Ibaraki, JAPAN
[3]Dep. of Telecommunication, Politeknik Negeri Semarang, Central Java, INDONESIA
Email: s.hidayat@aist.go.jp

## ABSTRACT

*Understanding object and its context are very important for robots when dealing with objects for completion of a mission. In this paper, an Affordance-based Ontology (ABO) is proposed for easy robot dealing with substantive and non-substantive objects. An ABO is a machine-understandable representation of objects and their relationships by what it's related to and how it's related. By using ABO, when dealing with a substantive object, robots can understand the representation of its object and its relation with other non-substantive objects. When the substantive object is not available, the robots have the understanding ability, in term of objects and their functions to select a non substantive object in order to complete the mission, such as giving raincoat or hat instead of getting stuck due to the unavailability of substantive object, e.g. umbrella. The experiment is done in the Ubiquitous Robotics Technology (u-RT) Space of National Institute of Advanced Industrial Science and Technology (AIST), Tsukuba, Japan.*

## 1. INTRODUCTION

Understanding the environment and its objects are a very important aspect in order for the robots to carry out its mission for serving human. A robot such as house holds service robots also have to understand and be capable to provide users' needs, e.g. if rain, robot will prepares umbrella for human. However, today's service robots' ability for dealing with everyday objects in dynamic-changing environment like a home is insufficient. Therefore, a robust and general engineering method for effectively and efficiently dealing with objects and users' needs are urgently needed.

To do so, firstly, the environment is structured by developing ubiquitous function for human life [1]. As a matter of fact, it is not so easy to build a distributed system in the daily living environment which has many kinds of sensors and actuators. To cope with these problems, the wireless network node named Ubiquitous Function Activation Module (UFAM) is developed which has highly versatile specifications to implement the ubiquitous space [2]. In the middleware platform, RT Middleware [3] is used to optimized the programming and on the most top layer Web Service [4] is applied. And also, the mobile manipulator is controlled using passive RFID tags implanted under the floor and make some robotic application systems for serving human [5]. The such environment is named as Ubiquitous Robotic Technology Space with ambient intelligence (u-RT Space for short) as shown in Fig.1[6].

In this paper, we proposed affordance based ontology (ABO) as machine-understandable representation of daily objects in the u-RT Space. The affordance concept is applied in the ontology as a kind of relationship between substantive object with other non-substantive object. Substantive object is an object which has an essential function belonging to the real nature or essential part of a thing. For example, an umbrella is substantive object which has essential function for protecting head from rain water or ultraviolet light. When an umbrella is applied as substantive object, others similarity-in-function objects, such as hat, cap, raincoat, newspaper, plastic bag, etc. are classified as non substantive objects. When robots deal with non-substantive objects, these kinds of objects will afford robots the same function as substantive object. By doing so, even the substantive object is not available robots can complete their mission by dealing with non-substantive object.

The rest of the paper is organized as follows. Section II describes related research. Section III describes affordance concept which is adopted into robotic field. Dealing objects in the physical layer is described in Section IV. The next section, Section V describes how the robots deal with the object representation in the semantic layer. Evaluations and discussions are described in Section VI. Finally we conclude this paper in Section VII

## 2. Related Research

There are several works in how robot system deals with an object. Most of them are applying a single action in response to single command [27]. As well as our daily live, living with robots in robotic environment such as a u-RT Space as shown in Fig.2, user's command may involve many different tasks, instead of single command, depending on situation, for example "*if raining outside, bring me umbrella*" task.

Furthermore, some of them have dealt using ontology [29], [30]. However, most researches have focused on users, object, and environment in providing everyday service, e.g. localization, task planning, etc. They have not focused in depth on relationships between objects and what they afford for robots (affordances). We tightly relate every object with its function based on affordance concept using ontology. Every object also has functional relationship (**hasFunction)** with other object or non-substantive object in order to compensate its availability, e.g. the object is no longer available. The proposed method will enable robots dealing with available objects which have same/similar function with the desired object to complete a mission/task even the desired object is not available

In previous works of our u-RT Space research group, has implemented two methods for dealing with objects and users' needs. First, to deal physical object by using physical information attached on the object such as RFID tags and then connected to the network through wireless communication node using UFAM. The robots get information about object's manipulation and its location such as how to grip a book from remote database. The successful implementation of this system is a librarian robotic system [7]. The second methods, to deal physical object which not connected to the network by applying visual mark [8] using QR code for object manipulation, e.g. how to grip the object. The above methods successfully dealt for manipulating object physically. However, a well prepared scenario to provide service in u-RT Space must be pre-program by user. In this scenario conditional expression must be change when object are added or reduce.

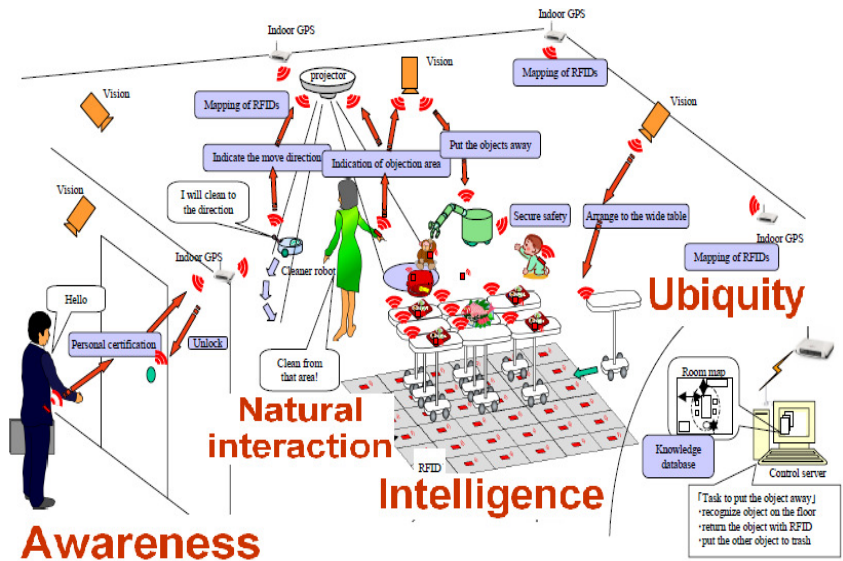

Figure 1. Concept of Ubiquitous Robots Technology (u-RT) Space, combining robot technology with Ambient Intelligence

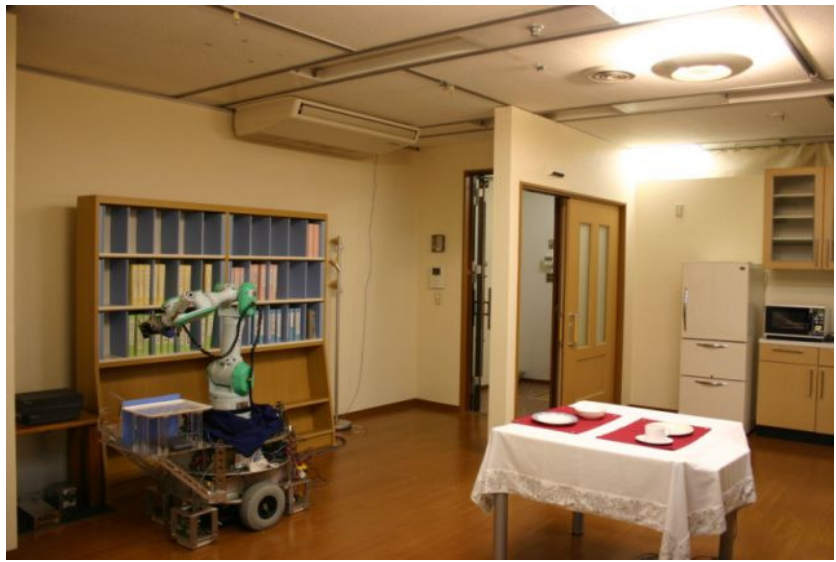

Figure 2. The test bed of Ubiquitous Robotics Technology (u-RT) Space

To overcome such condition, an object representation in semantic way, which consists of information about its relation with other objects and can be understood by machine, should be proposed and be integrated to the current u-RT system.

Dealing with object representation in semantic layer is quite different with dealing in physical layer. We applied object ontology by relating every daily objects in the environment which has similarity in functions based on affordance [9] concept.

The concept of affordance has been coined by J.J. Gibson on the ecological approach to visual perception and its link to action. Although he introduced the concept in psychology, it turned out to be elusive concept that influenced studies ranging from ecology, art science, industrial design, human-computer interaction and robotics [10-25].

## 3. AFFORDANCE CONCEPT

### 3.1. Basic Concept

The concept of affordance as shown in Fig.3, has been coined by J.J. Gibson [9] in his seminal work on the ecological approach to visual perception and its link to action. The concept of affordance is as he wrote:" The affordances of the environment are what it offers the animal, what it provides or furnishes, either for good or ill." [9]. In the context of ecological perception, visual perception would enable agents to experience in a direct way the opportunities for acting. However, Gibson remained unclear about both how this concept could be implemented in technical system and which representation to be used.

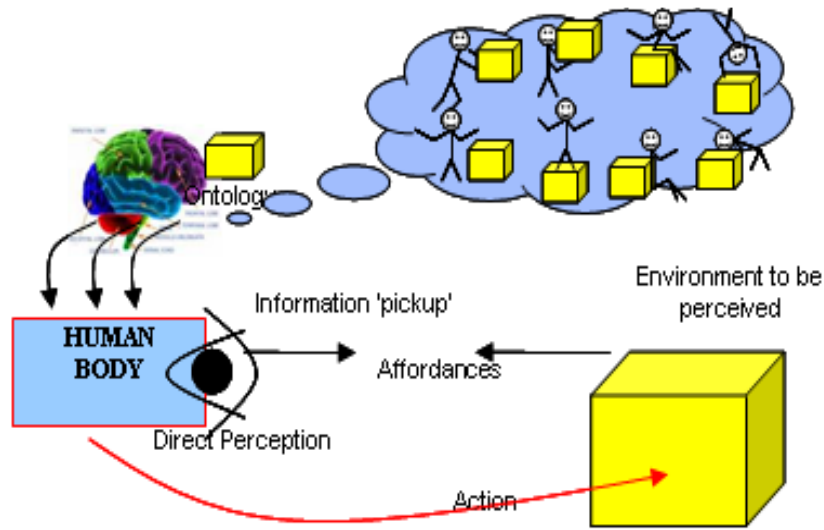

Figure 3. Dealing objects in human brain adopted as affordance concept.

### 3.2. Applying Affordances for Robotics

The concept of affordances which directly coupling perception to action from the object is highly related to autonomous robot control and influenced many studies in this field [11], [12], [13-15]. Starting from [12] she has developed and applied affordances approaches to mobile robots since last two decades. And recently, there are also other studies that exploit how affordances reflect to high-level processes such as tool-use [15], learning [16], [17] or decision-making [18].

How the relation between the concept of affordances and robotics and how robots learn affordances has started to be explicitly discussed by many roboticists. The co-relation between the theory of affordances and reactive/behavior-based robotics has already been pointed out in [13] and [14]. Stoytchev [15], [16] studied robot tools behavior as an approach to autonomous tool use, where the robot learns tools affordance to discovering tool-behavior pair that gives the desired effects. Fitzpatrick et al. [20] also study learning affordances in a developmental framework where a robot can learn what it can do with an object (e.g. rolling by tapping). Fritz et al. [17] demonstrated a system that learns to predict the lift-ability affordance. In this study, predictions are made based upon features of object regions, like color and shape description, which are extracted from the robot camera images. In both Stoytcheve's and Fitzpatrick's studies, no association between the visual features of the objects and their affordances, instead they used in both experiments the objects are differentiate using their colors only. How the robot learns the traversability affordance has recently been studied by Ugur et al. [18] and Kim et al. [21]. Contradicting with the last two researchers above, they used low level features, which are extracted from stereo vision or range image and used in learning and predicting of traversability affordance in unknown environment. Different from previous researcher, in [24] and [25] used imitation learning algorithm in order a humanoid robot learns object's affordances. Lopez used a probabilistic graphical model known as Bayesian networks to encode the dependencies between actions, object features and the effects of those actions. While Nishide used Recurrent Neural Network with Parametric Bias to predict object dynamics room visual images through active sensing experiences.

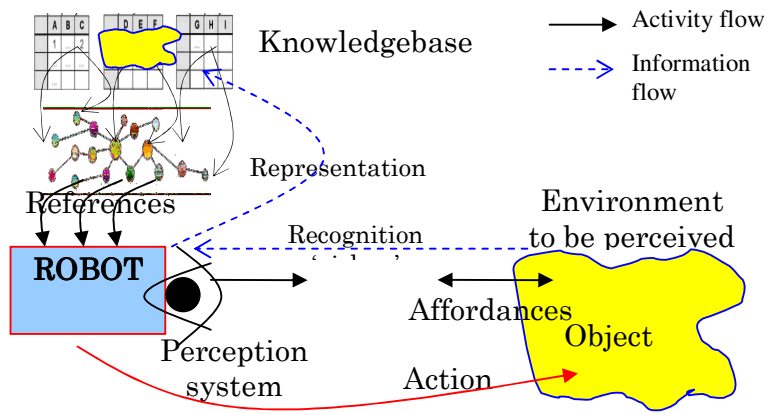

Figure 4. Applying affordance concept for robotics using ontology

In order to affordances for robotics make senses for human being, we are investigating semantic integration and ontology mapping and applied to ubiquitous robots to investigate what the environment, such as our daily objects afford for robots. Contradicting with other roboticists, we used physical landmark information attached on the objects as perceptual source for robots and process it semantically in order to obtain relevant affordances for appropriate/certain robots' task. To prove our concept, first we classify every physical object and the affordances *driver* into several classes, create ontology, define general properties, and make reasoning in order to verify the logical relation. Second, applying query engine to obtain the appropriate action which afforded by physical objects in certain situation and condition. And the last step is '*grounding*' the obtained affordance from text into context (robots and its environment). The proposed framework is depicted in Fig. 4.

This work has similar approach with [29] in using ontology-based knowledge for robots intelligence, however, we emphasize in implementation of affordances concept for robots using landmark tags, such as RFID tags, QR code.

## 4. DEALING IN PHYSICAL LAYER

This section describes some previous works in our group for dealing objects in the physical layer, which refers to how robots recognize the physical object and how its ability to manipulate it. The dealing method is describe in Fig.5.

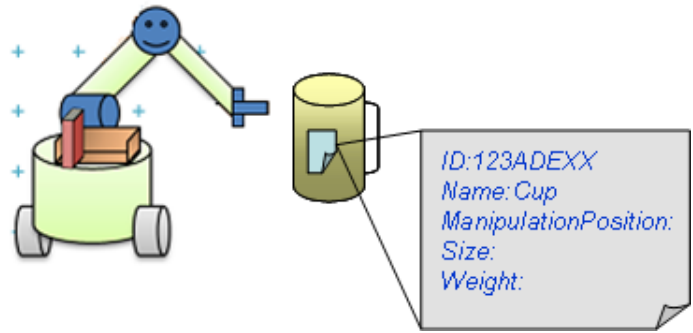

Figure 5.    A tagging information system for dealing with object in the u-RT Space

To deal with the physical objects, the physical hyperlinks have been developed using two kinds of RFID tags; active tags and passive tags as shown in Fig. 6. Each RFID tag has a native network address, which enables robots to access the object information through the network. As a result, this scheme allows robots to perceive space/location more easily and to handle objects more naturally, and realizes ambient intelligence.

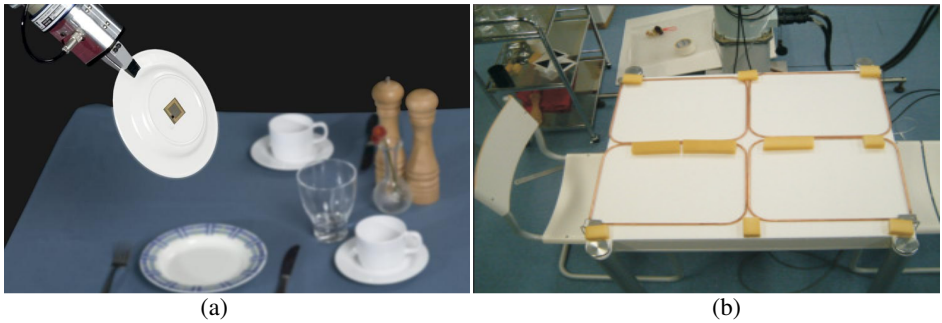

(a)                                                                                     (b)

Figure 6.    RFID tags implanted on the dishes (a) and the Tag Reader installed on the table (b)

Ubiquitous Functions Activation Module (UFAM) is developed as active tags for ubiquitous robots [1]. This device is shown in Fig.7. In general, UFAM can be embedded into every object in the smart environment. Using UFAM, the objects in the environment possess capability for information storage, processing, and communication, so they have a presence in both the physical and digital worlds. The robot can easily interact with these objects through digital interaction and physical interaction. For example when the UFAM tags included objects information are attached to some object, robot can easily recognize object name, size, color, how to use or manipulate it, etc.

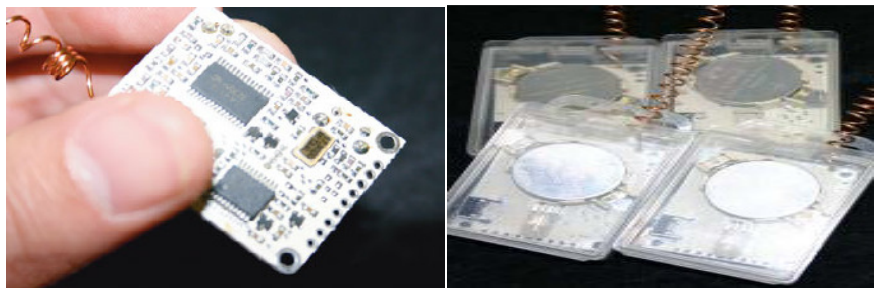

Figure 7.    Ubiquitous Functions Activation Module (UFAM) device

International Journal of Computer Science & Information Technology (IJCSIT) Vol 4, No 1, Feb 2012

A QR code (abbr. Quick Response code) is used for easy robots manipulation with object. In different way with previous method, Ohara et al [8]. proposed Coded landmark for Ubiquitous Environmnet (CLUE) as shown in Fig.8, as a visual mark using QR Code for easy robots manipulation with object. CLUE provides robots with information on the objects that are to be manipulated.

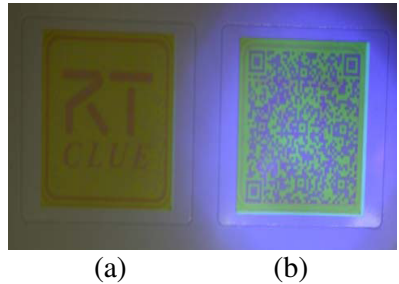

(a)         (b)

Figure 8.  Coded Landmark for Ubiquitous Environment (CLUE) based on QR code. (a) View under normal light source. (b) Under UV light source

In relation with design concept of everyday object, a universal handle as shown in Fig.9, which attached to the home appliances such as microwave oven, to help different types of robots' hand for objects manipulation, as well for human, has designed and successfully implemented in robotic system [8].

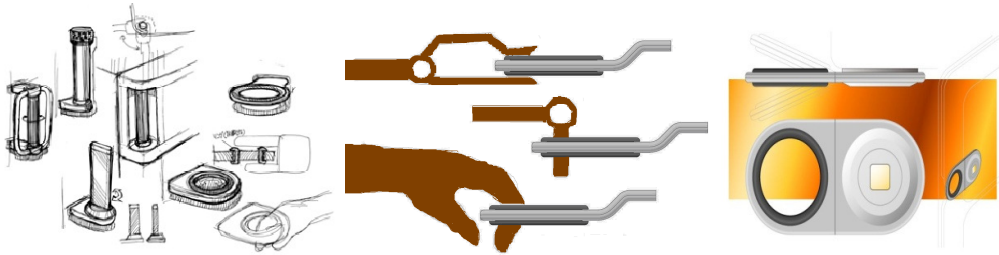

Figure 9.  Universal handle for dealing with different robot's hand and as well as human

Although the above works have shown highly effective results in dealing with physical object in the u-RT Space, they contain two remaining issues to be solved, i.e. objects' relationship and reasoning process. To cope with these problems, we proposed a dealing method in semantic layer for selection, instead of manipulation. The difference of both dealing methods is describe in Fig. 10.

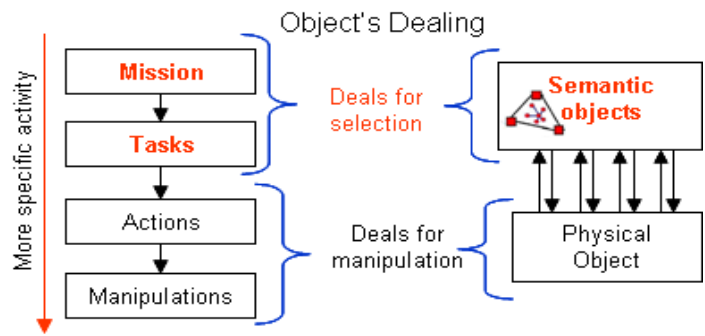

Figure 10. Dealing with semantic objects and physical object and its difference

## 5. DEALING IN SEMANTIC LAYER

### 5.1. *Addressed Problems*

There are two difficult conditions for robots to deal with object in u-RT Space as depicted in Fig.11. First, when the desired object is not longer available by some how, for example another family member suddenly took an umbrella just before robots doing so. In this scenario, robots cannot detect the umbrella longer, and the mission for serving human obviously failed.

The second condition, if the condition is opposite as the first one, means many objects which may support rainwater protector available there, such as raincoat, hat, cap, plastic bag, towel, old newspaper, etc. Even there are many objects which have same function as an umbrella and all objects can be recognized well by robots, robots without pre-programmed to do so, will not understand anything surrounding objects. By just knowing the object's properties, there is no way for new object inserted to the URT to be well understood by the robots for accomplishing a mission.

Due to dealing with objects in U-RT Space is limited by predefined rules; we need a new method by enabling reasoning process to cope dealing with surrounding objects without predefined rules. This will affect mission completion possibilities due to ability to use surrounding objects for supporting the mission.

### 5.2. *Concept's Terms and Definitions*

Due to many different terms and definitions referring to the same thing from different viewpoint, they should be defining in order to avoid ambiguity for understanding this concept.

The 'substantive'[1] word means an essential function belonging to the real nature or essential part of a thing. Therefore, we define the substantive objects are all objects in the u-RT Space. Substantive object refers to the object, which has substantive function. Every substantive object has substantive function. Some substantive objects have same substantive function.

Substantive function is an essential function belonging to the real nature or essential part of a thing. For example, an umbrella has a substantive function for protecting head or body part from rainwater or UV light. Hat, cap, raincoat, newspaper, etc. are objects, which also have functions same as umbrella for protecting body part from rainwater or UV light.

Non-substantive function refers to the other possibly functions of an object. For example, umbrella has non- substantive function such as stick, hand extension, tools, and few to name.

Dealing with physical object refers to the manipulation process such as how to grip, how to lift, etc. Dealing with semantic object refers to the object selection to deal with, e.g. umbrella, raincoat, newspaper, etc. Physical contact conducted with physical object in physical layer, while in semantic layer uses logical relationship to connect semantic object.

Physical layer or semantic layer refers to the abstraction layers related to physical object or semantic object representation. Semantic object is a representation of the physical object, both non/substantive objects.

---

[1] http://www.dictionary.com

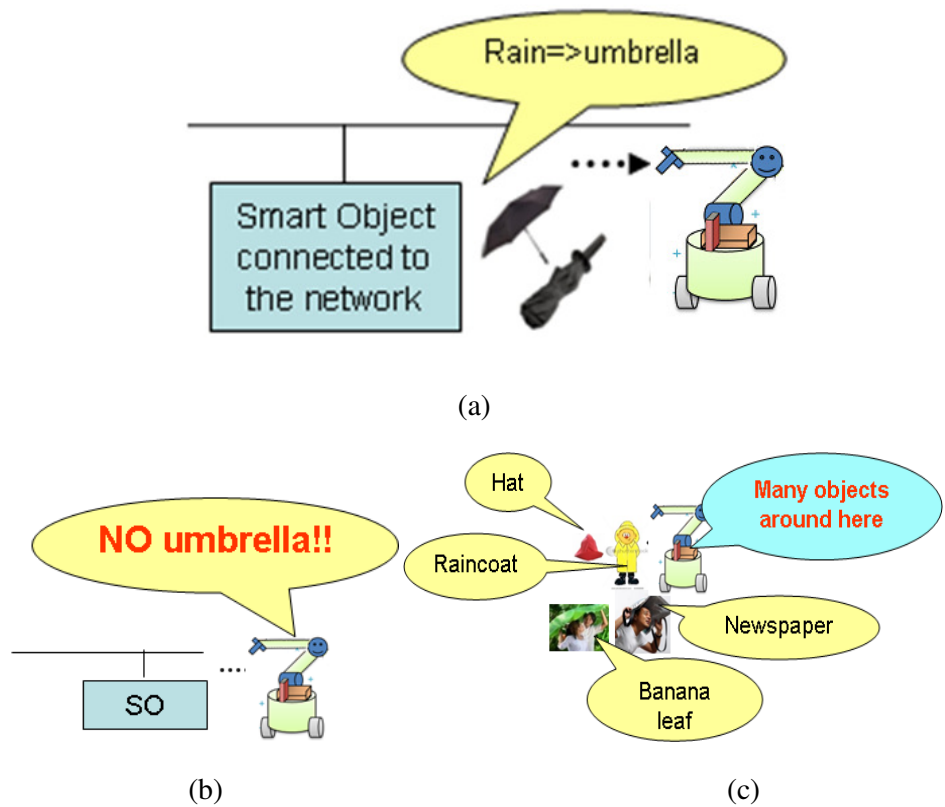

Figure 11. Some novel situations dealing with substantive objects in physical layer.(a) When substantive object available. (b) When the substantive object is not available, the system will lost its capability to perform such mission, e.g. providing service when its rain. (c) Same as (b) but there are many non-substantive object around robot

### 5.3. *Dealing in Semantic Layer Using Ontology*

Dealing in semantic layer is implemented using Semantic Web-related technology. To do so, first, the object is attached with RFID tags as shown in Fig. 12, to inform robot its capability, for example, a mug/cup affords water storage. Then, the object is represented in ontology using Semantic Web technology. Semantic Web technology enable categorization, communication and reasoning by providing standard protocols and languages for defining and sharing ontologies, using the Ontology Web Language (OWL). The result, a system integrating the robot physical's and sensory capabilities with high performance reasoning capabilities of the ontology inference engine, such as Racer, is a vast improvement over closed robotic system that are unable to novel situation.

The use of expressive ontologies in robotics allows for both feature-based and context-base categorization. Furthermore, ontologies enable these concepts and categorization to be shared with other robots.

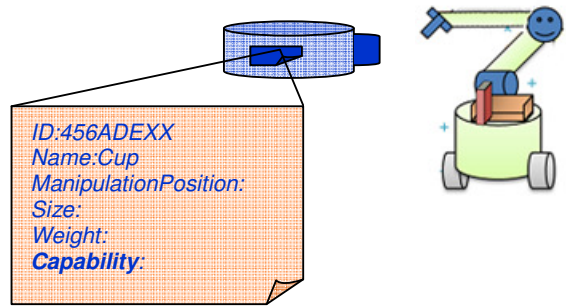

Figure 12. Adding object capability item and represented in the ontology

Using OWL properties, we can define abstract concept (such as the umbrella) from the elementarily concept. For example, as shown in Fig.13, a concept corresponding to **Umbrella** can have the necessary property **hasFunction** constrained to the concept **RainwaterProtector.** One can furthermore use OWL to define both sufficiency conditions: all objects with **hasFunction RainwaterProtector** are instances of **Umbrella.** Hence, we can assert other objects, such as raincoat, hat, cap, helmet, or even old newspaper as instances of **Umbrella**.

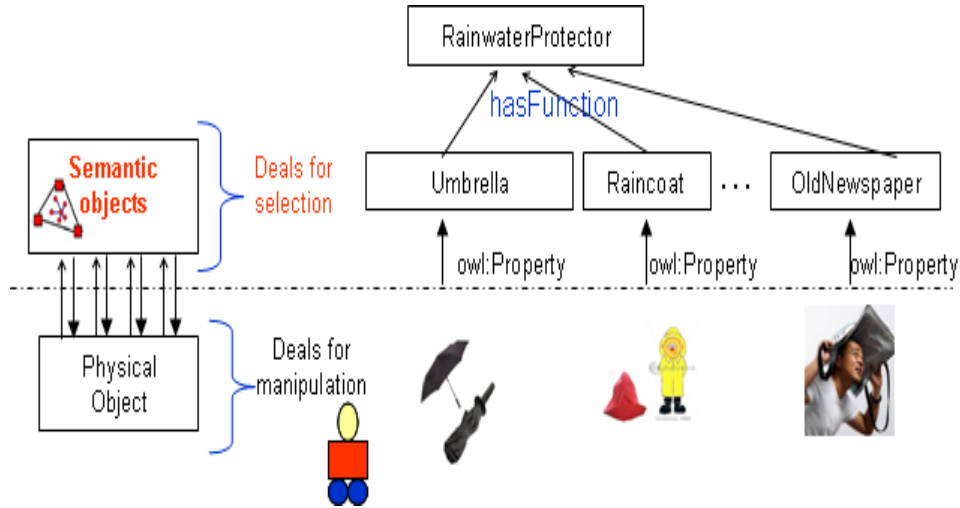

Figure 13. Dealing with semantic objects in semantic layer

Speaking in the physical layer, dealing with substantive object umbrella, which has substantive function as rainwater protector means if there is no umbrella, we/robots still can use/select non-substantive objects such as raincoat, hat, cap, or even old newspaper for providing rainwater protector. Hence, in affordance concept terminology, a non-substantive object, such old newspaper affords rainwater protector.

## 6. SYSTEM IMPLEMENTATION

The implementation of dealing methods with the physical objects has successfully implemented as we described in Section 4. In this section, we will describe implementation for service robot application dealing with object in semantic layer based on weather information as depicted in Fig. 14.

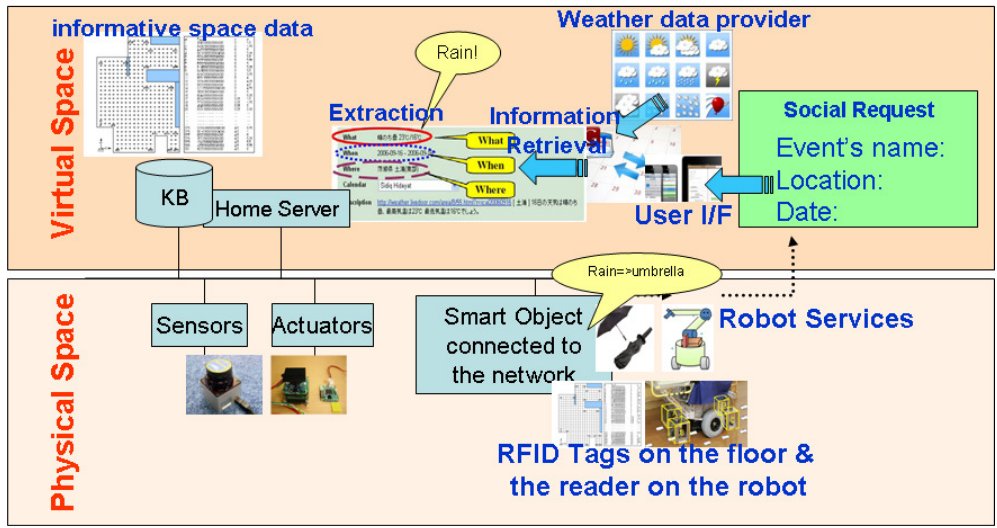

Figure 14. A service robot providing relevant object based on weather information.

If the mission is to provide user some services depending on weather forecast information, the sequence of information management controlled by the ubiquitous robot is simplified as follows:

1. The user writes event schedule on a handheld device and send it to his/her online calendar.

2. A middleware in smart environment detected an "event" from user's calendar and using feed reader read weather forecast for that day.

3. Before the user leaving home, a middleware system asked the robot to pick up an object based on weather information, e.g. if it rain take an umbrella.

4. The robot localizes the current position by reading the floor's tags and composes the path to navigate to the target position.

5. The robot navigates to the target position and compensates the path following error continuously using the information from the informative space.

6. If the robot arrives at the target position, the robot carries out the given task.

## 7. CONCLUSION

In this paper, the Affordances-based Ontology (ABO) is proposed and its prototype system is implemented. ABO exploits Semantic Web Services technology, a state of the art Web technology to provide interoperation between robots and objects in the u-RT Space environment.

ABO enables robots dealing with objects in semantic layer for selection as well as in physical layer for manipulation, by enabling reasoning process about capability of the environments. ABO is applied to solve the limitation of the u-RT Space in understanding the objects' relationships and its capability. By using ABO, the robots have the understanding ability (inferring new knowledge, in term of objects and their functions) to select a non-substantive object in order to complete the mission, rather then being stuck.

We have conducted this research in such ideal condition, e.g. Ambient Intelligence, where all objects are attached with tags such as RFID tags, in order to be well recognized and well localized by the system. For the future work, we need to prove this concept into natural environment such as common home environment.

For the comparison work in conventional everyday-service systems, usually they need a well prepared scenario to provide service [27]. In this scenario, conditional expression must be change when object are added or reduced. For example, in the cooking procedure proposed by Nakauchi [1], for cutting onion, the system cannot give solution if the user change the knife to other object, even the object also categorized as cutting instruments, such as paper cutter. By proposing and developing a system such as ABO, which allows robots to use surrounding objects in the natural environment, will be able to solve such limitation and give benefits for human life.

**Authors**

*Sidiq S. Hidayat* is a PhD student in Department of Intelligent Interaction Technology in University of Tsukuba, Japan. He is also a research student in Ubiquitous Function Research Group, a part of Intelligent Systems Research Institute (ISRI) at the National Institute of Advanced Industrial Science and Technology (AIST) Tsukuba, Japan. His research interests include the Semantic Web service, ontology, knowledge representation, commonsense reasoning, affordance concept and multiple agents systems. He is also a reviewer of IEEE Conference on Automation Science and Engineering (CASE) 2008 & 2009 and other conferences, member of @WAS and IECI.

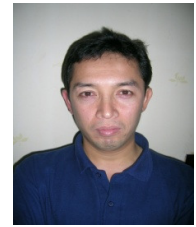

***Bong Keun Kim*** received the B.S. degree in mechanical and production engineering from Pusan National University, Busan, Korea, in 1994, and the M.S. and Ph.D. degrees in mechanical engineering from Pohang University of Science and Technology, Pohang, Korea, in 1996 and 2001, respectively. From 2001 to 2002, he was a Postdoctoral Fellow at the Automation Research Center, Pohang University of Science and Technology. From 2002 to 2003, he was a Postdoctoral Fellow at the University of California, Berkeley. From 2003 to 2005, he was a JSPS 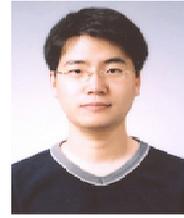 Postdoctoral Fellow at the ISRI, AIST Tsukuba, Japan. Since he joined AIST in 2005 as a Research Engineer, he has been working in the Ubiquitous Functions Research Group. His current research interests include control theory, robot middleware, ambient intelligence, ubiquitous robotics, and sensor network. He received best paper award in the Korea Intelligent Robot Conference, 2006 and silver prize in the 8th SAMSUNG HUMANTECH Thesis Award, 2002.

***Kohtaro Ohba*** is a group leader of the Dependable System Research Group, Intelligent Systems Research Institute at the AIST, Tsukuba, Japan. Prof. Ohba was born in Japan, in 1964, and received the B.S. degree, the M.S. degree, and the Ph.D. degree in mechanical engineering from the Tohoku University, Japan, in 1986, 1988, and 1991, respectively. After working at Tohoku University, he joined the Mechanical Engineering Laboratory (currently, the National Institute of Advanced Industrial Science and Technology), in 1997. From October of 1994 to June of 1996, 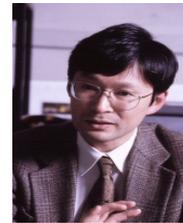 he worked in the School of Computer Science, Carnegie Mellon University, Pittsburgh, USA. His current research interests include ubiquitous robotics and ambient intelligence, object recognition, visualization, teleoperation, and human interface.